\documentclass{article}
\pdfoutput=1
\usepackage[dblblindworkshop,final]{neurips_2026}
\workshoptitle{3rd Workshop on Attributing Model Behavior at Scale: Data Attribution and Provenance}

\usepackage[utf8]{inputenc}
\usepackage[T1]{fontenc}
\usepackage{amsmath}
\usepackage{amssymb}
\usepackage{booktabs}
\usepackage{array}
\usepackage{graphicx}
\usepackage{hyperref}
\hypersetup{hidelinks}
\usepackage{microtype}
\usepackage{nicefrac}
\usepackage{tikz}
\usepackage{url}
\usepackage{xcolor}
\usetikzlibrary{arrows.meta,positioning}

\newcommand{\emj}[1]{\raisebox{-.12ex}{\includegraphics[height=1.05\fontcharht\font`X]{figures/emj-#1.pdf}}}
\newcommand{\affil}[1]{\textsuperscript{\normalfont #1}}

\title{Data Attribution Should Steer Pretraining}
\author{%
  \bfseries
  \href{https://openreview.net/profile?id=~Glenn_Matlin1}{Glenn Matlin}\affil{\emj{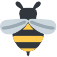}\,\emj{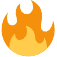}\,\emj{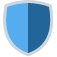}\,$*$} \quad
  \href{https://openreview.net/profile?id=~Chandreyi_Chakraborty1}{Chandreyi Chakraborty}\affil{\emj{bee}} \quad
  \href{https://openreview.net/profile?id=~Mohammed_Rehan_Parwani1}{Mohammed Rehan Parwani}\affil{\emj{bee}} \\[2pt]
  \bfseries
  \href{https://openreview.net/profile?id=~Saehee_Eom1}{Saehee Eom}\affil{\emj{bee}} \quad
  \href{https://openreview.net/profile?id=~Anthony_Wen-Ming_Zang1}{Anthony Wen-Ming Zang}\affil{\emj{bee}} \quad
  \href{https://openreview.net/profile?id=~Akhil_Theerthala1}{Akhil Theerthala}\affil{\emj{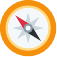}} \quad
  \href{https://openreview.net/profile?id=~Arjun_Chatterjee1}{Arjun Chatterjee}\affil{\emj{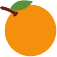}} \\[2pt]
  \bfseries
  \href{https://openreview.net/profile?id=~Tejas_Khandwekar1}{Tejas Khandwekar}\affil{\emj{bee}} \quad
  \href{https://openreview.net/profile?id=~Brandon_Michaels1}{Brandon Michaels}\affil{\emj{bee}} \quad
  \href{https://openreview.net/profile?id=~Alvin_Deng1}{Alvin Deng}\affil{\emj{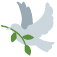}} \quad
  \href{https://openreview.net/profile?id=~Mark_Riedl1}{Mark Riedl}\affil{\emj{bee}} \\[4pt]
  \textnormal{\small
    \emj{bee}\,Georgia Institute of Technology\hspace{0.8em}%
    \emj{fire}\,MATS Program\hspace{0.8em}%
    \emj{shield}\,Georgia Tech AI Safety Initiative} \\
  \textnormal{\small
    \emj{tangerine}\,University of Illinois Urbana-Champaign\hspace{0.8em}%
    \emj{dove}\,EleutherAI\hspace{0.8em}%
    \emj{compass}\,Independent} \\[2pt]
  \textnormal{\footnotesize $^{*}$Corresponding author: \texttt{glenn@gatech.edu}}
}

\begin{document}

\maketitle

\begin{abstract}
Pretraining data curation decides what a model learns from, yet what makes a document worth learning from depends on what the model already knows. Influence-driven selectors already rank documents to improve one run; we argue that training-data attribution should steer pretraining by identifying testable data properties that become reusable curation criteria, evaluated on new data against cheaper selectors at matched total cost, discovery included. A document ranking is consumed by the run it was computed for, whereas an explained data property can be tested across corpora and checkpoints, especially where familiar quality signals diverge from measured usefulness. Because historical contribution, an understandable explanation, and future data value require distinct evidence, we specify a feedback loop that interprets attributed sources, tests candidate data changes, and states for five claim types what evidence each requires and what none establishes. We close with a falsifiable comparison against perplexity, classifier, and mixing baselines at equal total compute.
\end{abstract}

\section{Introduction}
\label{sec:position}

Pretraining data curation makes judgments about which text is worth learning from. Developers adjust mixture weights \citep{chen2025aioli}, filter subcorpora \citep{penedo2024fineweb}, reorder curricula \citep{zhang2026beyond}, and compare curation strategies in controlled testbeds \citep{li2024datacomplm}. Attribution adds a way to investigate how training instances contribute to a model's behavior, under a specified account of influence \citep{koh2017understanding,grosse2023studying,park2023trak}.

Language models evolve dynamically as optimization interacts with data mixtures, capacity, and objectives \citep{biderman2026fix}. Data that helped form an ability may become redundant; other material becomes useful once prerequisites develop. Influence-driven curricula \citep{babylm2025influence} and model-aware selection and mixing \citep{engstrom2024dsdm,yu2024mates,wang2025tikmixdatainfluencedynamic} already act on attribution during pretraining, with gains and losses across tested policies. The broader opportunity is to learn which data properties support which capabilities under which training conditions, and hold those decisions to matching evidence.

\textbf{Our position is that training-data attribution should steer pretraining by identifying testable data properties that become reusable curation criteria, evaluated on new data against cheaper selectors at matched total cost, discovery included.} Influence rankings are consumed by the run they were computed for; a property of quality, content, or structure that explains a gain can be tested beyond that run and compiled into a selector that needs no fresh attribution. We call the study of these links capability provenance: connecting a target ability to the data and training conditions that shaped it \citep{matlin2026capability}.

Prior provenance work traced abilities in trained models; prospective steering asks what evidence those links require before changing a training mix, and pairs developer-facing interpretability \citep{doshivelez2017rigorous,nguyen2024towards} with actionable decision utility \citep{orgad2026actionable}. We report no new experiments because this claim concerns what should count as evidence for a class of curation decisions, which no single run settles. Section~\ref{sec:research-agenda} sets out the matched-cost comparison the field should run in open testbeds, and Section~\ref{sec:alternative-views} states what would refute the position.

\section{Historical Contribution and Future Data Value}
\label{sec:data-value}

Our scope is \emph{contributive attribution}, how training data shapes behavior, rather than corroborative attribution of sources that support an output \citep{worledge2023unifying}. Within contributive attribution, removing a document, reweighting it, and replacing a mixture ask distinct counterfactuals \citep{bae2022if}: for a metric $f$ of trained parameters, removing $z$ asks $f(\theta_{-z})-f(\theta)$ over training randomness, reweighting asks $\partial f(\theta_{\varepsilon})/\partial\varepsilon$ at $\varepsilon=0$ for weight $1+\varepsilon$, and mixture replacement asks $f(\theta_{w'})-f(\theta_{w})$, so an estimate for one action does not license another.

Data that helped form a capability may add little once the model has learned from it. Conversely, data with little contribution so far may become useful after prerequisites develop \citep{chen2023skillitdatadrivenskillsframework,dang2026capacity}, and data introduced at another stage can have a different effect \citep{liu2026midtrainingbridgespretrainingposttraining}. These possibilities defeat a rule that the most influential past examples should receive the most future weight. Online selection for what is learnable and not yet learnt \citep{mindermann2022prioritized}, influence models of evolving preferences \citep{yu2024mates}, and optimal-control selection \citep{gu2025dataselectionoptimalcontrol} act on this premise. The object of study is conditional data value $\Delta_f(a \mid \theta_t, w, B_t)$: the change in metric $f$ from data change $a$ at checkpoint $\theta_t$, mixture $w$, and remaining budget $B_t$.

Because conditional value shifts as prerequisites form, the training path carries signal a final checkpoint erases. Gradient tracing \citep{pruthi2020estimating} and stagewise attribution \citep{lee2026influence} track this path, and task-agnostic analysis finds shifting contributions toward final parameters \citep{nishida2026measuring}, though parameter progress alone does not prove target capability. Since fixed-target attribution runs online at low overhead \citep{wang2025inrunshapley} while influence shifts across stages \citep{wang2024capturingtemporaldependencetraining}, the question is which decisions trajectory estimates resolve that an endpoint cannot.

Content quality and contribution also come apart: influential text can teach a shortcut, while filtered text can supply useful structure. On English C4 and Pile web text for 125M--1.7B models, reference-corpus and off-the-shelf embedding similarity fail to beat random selection or datamodels \citep{engstrom2024dsdm,sam2025analyzingsimilaritymetricsdata}; on RedPajama-V2 and multi-recipe pools across 125M--7B models, static quality classifiers and proxy rankings diverge from reference loss and token utility \citep{saada2026removingnoisefindinggold,lin2024rho1,wang2026smalltrainingruns}; and in multi-epoch runs up to 7B--9B parameters on token-limited Common Crawl and C4 pools, unfiltered text overtakes filtered subsets once repetition exhausts the filtered pool \citep{muennighoff2023scaling,goyal2024scaling,mohri2026bitterlessondatafiltering}. The open question is which data properties explain these regime-dependent disagreements and whether attribution can surface them before a run.

\section{Capability Provenance as a Control Loop}
\label{sec:control-loop}

\begin{figure}[t]
  \centering
  \definecolor{loopblue}{HTML}{0072B2}
  \definecolor{looporange}{HTML}{E69F00}
  \definecolor{loopgray}{HTML}{666666}
  \tikzset{
    loopbox/.style={draw=loopblue, fill=loopblue!9, rounded corners=2pt,
      minimum height=1.1cm, text width=2.65cm, align=center,
      font=\small\bfseries, inner sep=4pt, line width=0.7pt},
    validatebox/.style={loopbox, draw=looporange!80!black,
      fill=looporange!15, minimum height=1.42cm, text width=3.05cm},
    stopbox/.style={draw=loopgray, fill=black!3, dashed,
      rounded corners=2pt, minimum height=1.1cm, text width=2cm,
      align=center, font=\footnotesize\bfseries, inner sep=4pt,
      line width=0.7pt},
    looparrow/.style={-{Latex[length=2mm]}, line width=0.8pt},
    stoparrow/.style={looparrow, dashed, draw=loopgray}
  }
  \begin{tikzpicture}[node distance=0.72cm and 0.42cm]
    \node[loopbox] (observe) {Observe\\learning};
    \node[loopbox, right=of observe] (attribute) {Attribute and\\interpret sources};
    \node[validatebox, right=of attribute] (validate) {Assess evidence\\for this decision};
    \node[stopbox, right=0.34cm of validate] (stop) {Revise the\\hypothesis\\or narrow claim};
    \node[loopbox, below=1.05cm of validate] (change) {Test a data\\policy change};
    \node[loopbox] (update) at (observe |- change) {Update policy\\and criteria};
    \path (update) -- (change) node[midway, loopbox] (measure)
      {Measure target\\and side effects};
    \draw[looparrow] (observe) -- (attribute);
    \draw[looparrow] (attribute) -- (validate);
    \draw[looparrow] (validate) -- node[right, font=\scriptsize, align=left]
      {choose a\\bounded test} (change);
    \draw[looparrow] (change) -- (measure);
    \draw[looparrow] (measure) -- (update);
    \draw[looparrow] (update.north) -- node[left, font=\scriptsize] {next run}
      (observe.south);
    \draw[stoparrow] (validate) -- (stop);
  \end{tikzpicture}
  \caption{The attribution-to-steering control loop. Evidence supports a test proportionate to the decision's cost and uncertainty; outcomes inform later data policies and curation criteria, or send an unsupported interpretation back along the dashed path.}
  \label{fig:control-loop}
\end{figure}
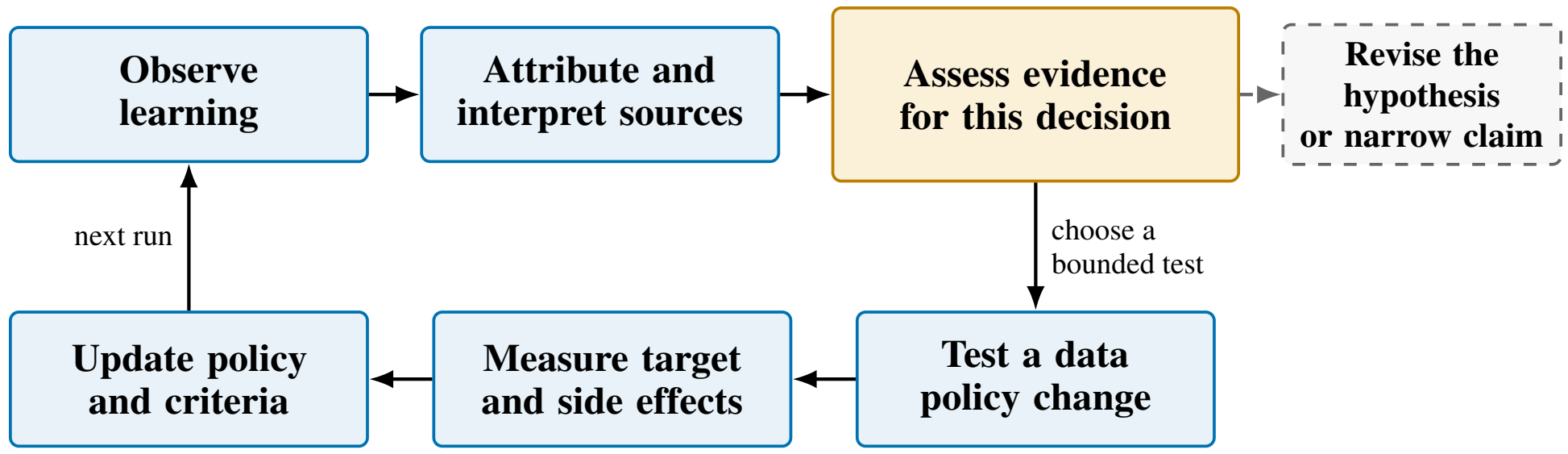

\paragraph{Observe, attribute, and interpret.}
In the loop of Figure~\ref{fig:control-loop}, developers track capability and safety metrics across checkpoints and match the unit of attribution to the unit of intervention. Broad buckets can conceal opposing effects, and summed document scores do not equal a group effect \citep{guu2023simfluence,yu2025groupmates,heo2026interactionawareinfluencefunctionsgroup}. A readable label only proposes a hypothesis, since discourse structure, source identity, and lexical overlap can drive the same score; when explanations remain entangled at the evidentiary check (Section~\ref{sec:evidence-contract}), the loop takes the dashed revision path.

\paragraph{Test a data change and update criteria.}
A bounded trial, such as an end-of-training ablation \citep{blakeney2024doesdatasparkjoy}, can test an uncertain attribution before a larger run. Every data change evaluates its target capability, adjacent capabilities, safety, and general language modeling as a joint outcome vector, where a gain on one metric cannot offset an unacceptable regression on another. Findings may also suggest a revised criterion or selector, as expensive judgments are distilled into cheap raters \citep{wettig2024qurating,yu2024mates} and tested on new data at matched cost.

\noindent\textit{Published example.} \citet{coalson2025ifguide} trace toxicity in Pythia-160M trained on 1B OpenWebText tokens using EK-FAC influence on RealToxicityPrompts. Standard document-level attribution fails on intervention: dropping the top 10\% of documents cuts toxicity by $\le$10\%, whereas dropping 50\% cuts toxicity by 33\% but degrades perplexity by 21\% and LAMBADA by 13\%. Following Figure~\ref{fig:control-loop}'s dashed path separates three conflated drivers in those document scores: shared function words, isolated outlier tokens in benign text, and local toxic context. Subtracting non-toxic query gradients and penalizing dense 99th-percentile token clusters (2\% of tokens in a one-token window) during pretraining cuts held-out explicit and implicit toxicity 4--10$\times$ up to 1B parameters within baseline fluency bounds, and token selections from the 160M proxy transfer across model scales on the same pretraining corpus.

\section{A Ladder of Information for Data Decisions}
\label{sec:evidence-contract}

Table~\ref{tab:contract} separates five evidentiary questions (historical contribution, explanatory property, out-of-sample prediction \citep{ilyas2022datamodels,park2023trak}, controlled intervention, and learned mechanism) alongside what each cannot establish. Because these rows answer distinct questions rather than forming a mandatory sequence, a selective intervention can precede mechanistic understanding, even though a successful action still leaves its motivating explanation unconfirmed.

\begin{table}[ht]
  \caption{Different attribution claims require different evidence and license different conclusions.}
  \label{tab:contract}
  \centering
  \footnotesize
  \renewcommand{\arraystretch}{1.05}
  \begin{tabular}{@{}>{\raggedright\arraybackslash}p{0.22\linewidth}>{\raggedright\arraybackslash}p{0.35\linewidth}>{\raggedright\arraybackslash}p{0.35\linewidth}@{}}
    \toprule
    \textbf{Question} & \textbf{Relevant evidence} & \textbf{What it does not establish} \\
    \midrule
    \textbf{1.}~Which data contributed? & Attribution estimates under a stated target and counterfactual. & That influential data will be useful in a later training decision. \\
    \textbf{2.}~What might explain the contribution? & Inspection of content, lexical and structural patterns, and competing explanations. & That an understandable pattern faithfully explains the model's dependence on the data. \\
    \textbf{3.}~Does the relationship predict unseen outcomes? & Predictions evaluated on evidence distinct from that used to construct them. & That a forecast of attribution scores predicts the effect of changing training data. \\
    \textbf{4.}~Does the candidate data change help? & A comparison isolating the intervention across target and off-target effects. & That the motivating explanation was right, or that the policy transfers. \\
    \textbf{5.}~How does the data shape computation? & A separately validated account linking data, representations, and behavior. & That this account identifies the best future curation policy. \\
    \bottomrule
  \end{tabular}
\end{table}

Walking \citet{coalson2025ifguide} through Table~\ref{tab:contract} clarifies what that study licenses for curation: failed document deletion exposed a granularity mismatch in row~1, differential token-density scoring isolated a candidate property in row~2, and pretraining suppression plus layer probing supplied scoped evidence for rows~4--5. Because the cross-scale test reuses fixed token indices within the same OpenWebText slice, however, it leaves open whether the extracted property transfers to a held-out corpus or beats cheap token classifiers \citep{rathi2026shapingcapabilitiestokenleveldata} once EK-FAC discovery compute is charged. Across all five rows, \emph{use}, \emph{identity}, \emph{interpretation}, \emph{robustness}, and \emph{causal support} qualify each claim (Appendix~\ref{app:claim-ladder}).

\section{Empirical Precedents and Measurement Limits}
\label{sec:foothold}

Fact tracing finds influential examples that support outputs via relation and entity priors without stating the queried fact \citep{chang2025scalable}. Tracing studies contrast memorization-like support for knowledge with generalization-like support for reasoning \citep{wang2025generalizationvsmemorizationtracing}, capability-provenance work separates support profiles across social and technical probes via targeted unlearning \citep{matlin2026capability}, and influence selects instruction data for targeted skills \citep{xia2024less}. These studies surface unexpected dependencies while exposing two measurement limits.

On the target side, a benchmark score samples one capability facet rather than defining it, and near-chance accuracy limits attribution. On the intervention side, unlearning is an imperfect proxy because suppressed behavior can recover from independent facts \citep{deeb2025unlearninginfo,lucki2025adversarial} while inducing neighbor knowledge holes \citep{ko2025probing}; neither single-benchmark attribution nor an unlearning screen validates prospective upweighting.

\section{A Research Agenda for Pretraining Decisions}
\label{sec:research-agenda}

\paragraph{Benchmark reusable criteria at matched total cost.}
Open testbeds such as DataComp-LM, DATE-LM, and DataDecide \citep{li2024datacomplm,jiao2025date,magnusson2025datadecide} can test whether attribution-derived criteria transfer to held-out corpora better than domain reweighting and regression mixing \citep{xie2023doremi,liu2025regmixdatamixtureregression}, importance resampling \citep{xie2023dsir}, online mixing \citep{albalak2023efficient}, model-aware and distilled selection \citep{yu2024mates,shum2025predictivedataselectiondata}, and direct curation \citep{jiang2024adaptivedataoptimizationdynamic} under a named estimator. Total cost sums compute for indexing, scoring, ablations, and training plus capped inspection; charging discovery grants baselines the same budget for extra tokens or mixture sweeps on the held-out corpus (Appendix~\ref{app:claim-ladder}, Table~\ref{tab:protocol}).

\paragraph{Instrument outcomes across stages and held-out formats.}
Section~\ref{sec:foothold}'s limits govern scoring: evaluations must hold out task formats disjoint from attribution queries under pre-declared adjacent-capability, safety, and perplexity bounds. Because unlearned behavior can recover from independent facts, unlearning only screens candidates, whereas retraining or continual-pretraining branches across checkpoints test how value shifts.

\paragraph{Separate failed explanations from estimator error.}
A failed held-out run may reflect either a false data-value hypothesis or an unfaithful influence approximation. Comparing optimizer-aware trajectory estimates \citep{deng2026faithful} with small-model retraining isolates estimator error before discarding a property, and logging the hypothesis, checkpoint, estimator, and outcome in provenance records \citep{gebru2021datasheets} keeps negative findings cumulative.

\section{Alternative Views, Falsification, and Scope}
\label{sec:alternative-views}

Direct curation ablations and cheaper selectors such as perplexity pruning, perplexity correlations, or synthetic data probes may resolve the decision \citep{ankner2025perplexed,thrush2025improvingpretrainingdatausing,li2024datacomplm,wang2026positionletsdevelopdata}. Because current estimators carry query sensitivity and non-convex approximation error \citep{basu2021influence,bae2022if}, raw scores require separating controls (Section~\ref{sec:evidence-contract}); attribution earns its cost when opposing effects cancel in coarse buckets or competing explanations prescribe different rules, complementing structural-alignment accounts on the training corpus \citep{lehalleur2025youarewhatyoueat}. The position fails if criteria derived from attribution do not beat perplexity-based or classifier selectors on held-out corpora at equal total compute, discovery included.

Capability provenance is also dual-use: source maps that aid bias and safety audits can guide targeted poisoning or concentrate harmful skills. Section~\ref{sec:control-loop} therefore pairs interventions with safety bounds, and empirical contribution never confers permission to train on contested data.

\section{Conclusion}
\label{sec:conclusion}

Training-data attribution should be evaluated by its utility for pretraining decisions, and its distinctive utility is explanation. The most influential examples in a model's past need not be the ones to show it next; learning why is what attribution adds when its criteria win at matched cost.

\clearpage
\begin{ack}
We thank Bhavya Shah and Isaac Song for discussions, early drafting, and feedback on the manuscript, and we thank the ATTRIB 2026 reviewers and program chairs for constructive comments. We also thank the ML Alignment \& Theory Scholars (MATS) Program, the Berkeley Existential Risk Initiative (BERI), and the Georgia Tech AI Safety Initiative for research and program support. The authors declare no competing financial interests.
\end{ack}

\bibliographystyle{plainnat}
\bibliography{references,references-vision}

\begin{thebibliography}{75}
\providecommand{\natexlab}[1]{#1}
\providecommand{\url}[1]{\texttt{#1}}
\expandafter\ifx\csname urlstyle\endcsname\relax
  \providecommand{\doi}[1]{doi: #1}\else
  \providecommand{\doi}{doi: \begingroup \urlstyle{rm}\Url}\fi

\bibitem[Aky{\"u}rek et~al.(2022)Aky{\"u}rek, Bolukbasi, Liu, Xiong, Tenney,
  Andreas, and Guu]{Akyurek_2022}
Ekin Aky{\"u}rek, Tolga Bolukbasi, Frederick Liu, Binbin Xiong, Ian Tenney,
  Jacob Andreas, and Kelvin Guu.
\newblock Towards tracing knowledge in language models back to the training
  data.
\newblock In \emph{Findings of the Association for Computational Linguistics:
  EMNLP 2022}, pages 2429--2446. Association for Computational Linguistics,
  2022.
\newblock \doi{10.18653/v1/2022.findings-emnlp.180}.
\newblock URL \url{https://aclanthology.org/2022.findings-emnlp.180/}.

\bibitem[Albalak et~al.(2023)Albalak, Pan, Raffel, and
  Wang]{albalak2023efficient}
Alon Albalak, Liangming Pan, Colin Raffel, and William~Yang Wang.
\newblock Efficient online data mixing for language model pre-training.
\newblock arXiv preprint arXiv:2312.02406, 2023.
\newblock URL \url{https://arxiv.org/abs/2312.02406}.

\bibitem[Ankner et~al.(2025)Ankner, Blakeney, Sreenivasan, Marion, Leavitt, and
  Paul]{ankner2025perplexed}
Zachary Ankner, Cody Blakeney, Kartik Sreenivasan, Max~M. Marion, Matthew
  Leavitt, and Mansheej Paul.
\newblock Perplexed by perplexity: Perplexity-based data pruning with small
  reference models.
\newblock In \emph{International Conference on Learning Representations}, pages
  100615--100634, 2025.
\newblock URL \url{https://arxiv.org/abs/2405.20541}.

\bibitem[Bae et~al.(2022)Bae, Ng, Lo, Ghassemi, and Grosse]{bae2022if}
Juhan Bae, Nathan Ng, Alston Lo, Marzyeh Ghassemi, and Roger~B. Grosse.
\newblock If influence functions are the answer, then what is the question?
\newblock In \emph{Advances in Neural Information Processing Systems},
  volume~35, pages 17953--17967, 2022.
\newblock URL \url{https://arxiv.org/abs/2209.05364}.

\bibitem[Bae et~al.(2024)Bae, Lin, Lorraine, and
  Grosse]{bae2024trainingdataattributionapproximate}
Juhan Bae, Wu~Lin, Jonathan Lorraine, and Roger Grosse.
\newblock Training data attribution via approximate unrolling.
\newblock In \emph{Advances in Neural Information Processing Systems},
  volume~37, pages 66647--66686, 2024.
\newblock \doi{10.52202/079017-2129}.
\newblock URL \url{https://arxiv.org/abs/2405.12186}.

\bibitem[Basu et~al.(2021)Basu, Pope, and Feizi]{basu2021influence}
Samyadeep Basu, Philip Pope, and Soheil Feizi.
\newblock Influence functions in deep learning are fragile.
\newblock In \emph{International Conference on Learning Representations}, 2021.
\newblock URL \url{https://arxiv.org/abs/2006.14651}.

\bibitem[Biderman et~al.(2026)Biderman, Khan, Mireshghallah, Arnett, Barez, and
  Saphra]{biderman2026fix}
Stella Biderman, Mohammad~Aflah Khan, Niloofar Mireshghallah, Catherine Arnett,
  Fazl Barez, and Naomi Saphra.
\newblock Position: Don't just ``fix it in post'': A science of {AI} must study
  training dynamics.
\newblock In \emph{Proceedings of the 43rd International Conference on Machine
  Learning}, 2026.
\newblock URL \url{https://arxiv.org/abs/2606.06533}.

\bibitem[Blakeney et~al.(2024)Blakeney, Paul, Larsen, Owen, and
  Frankle]{blakeney2024doesdatasparkjoy}
Cody Blakeney, Mansheej Paul, Brett~W. Larsen, Sean Owen, and Jonathan Frankle.
\newblock Does your data spark joy? {P}erformance gains from domain upsampling
  at the end of training.
\newblock arXiv preprint arXiv:2406.03476, 2024.
\newblock URL \url{https://arxiv.org/abs/2406.03476}.

\bibitem[Chang et~al.(2025{\natexlab{a}})Chang, Li, Huber, Vogeti, Kant, Shi,
  and Chandra]{chang2026automixercheckpointartifactsautomatic}
Ernie Chang, Yang Li, Patrick Huber, Vish Vogeti, David Kant, Yangyang Shi, and
  Vikas Chandra.
\newblock {AutoMixer}: Checkpoint artifacts as automatic data mixers.
\newblock In \emph{Proceedings of the 63rd Annual Meeting of the Association
  for Computational Linguistics (Volume 1: Long Papers)}, 2025{\natexlab{a}}.
\newblock URL \url{https://arxiv.org/abs/2506.21910}.

\bibitem[Chang et~al.(2024)Chang, Park, Ye, Yang, Seo, Chang, and
  Seo]{chang2024largelanguagemodelsacquire}
Hoyeon Chang, Jinho Park, Seonghyeon Ye, Sohee Yang, Youngkyung Seo, Du-Seong
  Chang, and Minjoon Seo.
\newblock How do large language models acquire factual knowledge during
  pretraining?
\newblock In \emph{Advances in Neural Information Processing Systems}, 2024.
\newblock URL \url{https://arxiv.org/abs/2406.11813}.

\bibitem[Chang et~al.(2025{\natexlab{b}})Chang, Rajagopal, Bolukbasi, Dixon,
  and Tenney]{chang2025scalable}
Tyler~A. Chang, Dheeraj Rajagopal, Tolga Bolukbasi, Lucas Dixon, and Ian
  Tenney.
\newblock Scalable influence and fact tracing for large language model
  pretraining.
\newblock In \emph{International Conference on Learning Representations},
  2025{\natexlab{b}}.
\newblock URL \url{https://arxiv.org/abs/2410.17413}.

\bibitem[Chen et~al.(2026)Chen, Luo, and Pan]{chen2026mechanistic}
Jianhui Chen, Yuzhang Luo, and Liangming Pan.
\newblock Mechanistic data attribution: Tracing the training origins of
  interpretable {LLM} units.
\newblock In \emph{Proceedings of the 43rd International Conference on Machine
  Learning}, 2026.
\newblock URL \url{https://arxiv.org/abs/2601.21996}.

\bibitem[Chen et~al.(2023)Chen, Roberts, Bhatia, Wang, Zhang, Sala, and
  R{\'e}]{chen2023skillitdatadrivenskillsframework}
Mayee~F. Chen, Nicholas Roberts, Kush Bhatia, Jue Wang, Ce~Zhang, Frederic
  Sala, and Christopher R{\'e}.
\newblock {Skill-It!} {A} data-driven skills framework for understanding and
  training language models.
\newblock In \emph{Advances in Neural Information Processing Systems}, 2023.
\newblock URL \url{https://arxiv.org/abs/2307.14430}.

\bibitem[Chen et~al.(2025)Chen, Hu, Lourie, Cho, and R{\'e}]{chen2025aioli}
Mayee~F. Chen, Michael~Y. Hu, Nicholas Lourie, Kyunghyun Cho, and Christopher
  R{\'e}.
\newblock Aioli: A unified optimization framework for language model data
  mixing.
\newblock In \emph{International Conference on Learning Representations}, pages
  46089--46132, 2025.
\newblock URL \url{https://arxiv.org/abs/2411.05735}.

\bibitem[Choenni et~al.(2023)Choenni, Garrette, and Shutova]{Choenni_2023}
Rochelle Choenni, Dan Garrette, and Ekaterina Shutova.
\newblock How do languages influence each other? {S}tudying cross-lingual data
  sharing during {LM} fine-tuning.
\newblock In \emph{Proceedings of the 2023 Conference on Empirical Methods in
  Natural Language Processing}, pages 13244--13257. Association for
  Computational Linguistics, 2023.
\newblock \doi{10.18653/v1/2023.emnlp-main.818}.
\newblock URL \url{https://aclanthology.org/2023.emnlp-main.818/}.

\bibitem[Coalson et~al.(2025)Coalson, Bae, Carlini, and
  Hong]{coalson2025ifguide}
Zachary Coalson, Juhan Bae, Nicholas Carlini, and Sanghyun Hong.
\newblock {IF-GUIDE}: Influence function-guided detoxification of {LLM}s.
\newblock In \emph{Advances in Neural Information Processing Systems}, 2025.
\newblock URL \url{https://arxiv.org/abs/2506.01790}.

\bibitem[Dang et~al.(2026)Dang, Just, and Jia]{dang2026capacity}
Cuong Dang, Hoang~Anh Just, and Ruoxi Jia.
\newblock Capacity-dependent effects of data selection for reasoning.
\newblock In \emph{Conference on Language Modeling}, 2026.
\newblock URL \url{https://arxiv.org/abs/2608.13721}.

\bibitem[Deeb and Roger(2025)]{deeb2025unlearninginfo}
Aghyad Deeb and Fabien Roger.
\newblock Do unlearning methods remove information from language model weights?
\newblock arXiv preprint arXiv:2410.08827, 2025.
\newblock URL \url{https://arxiv.org/abs/2410.08827}.

\bibitem[Deng et~al.(2026)Deng, Hu, Jin, Lu, Wang, Zhang, and
  Ma]{deng2026faithful}
Junwei Deng, Pingbang Hu, Suliang Jin, Hao Lu, Jiachen~T. Wang, Shichang Zhang,
  and Jiaqi~W. Ma.
\newblock How faithful is trajectory-based data attribution? {E}rror sources,
  remedies, and practical guidelines.
\newblock arXiv preprint arXiv:2605.18814, 2026.
\newblock URL \url{https://arxiv.org/abs/2605.18814}.

\bibitem[Doshi-Velez and Kim(2017)]{doshivelez2017rigorous}
Finale Doshi-Velez and Been Kim.
\newblock Towards a rigorous science of interpretable machine learning.
\newblock arXiv preprint arXiv:1702.08608, 2017.
\newblock URL \url{https://arxiv.org/abs/1702.08608}.

\bibitem[Engstrom et~al.(2024)Engstrom, Feldmann, and Madry]{engstrom2024dsdm}
Logan Engstrom, Axel Feldmann, and Aleksander Madry.
\newblock {DsDm}: Model-aware dataset selection with datamodels.
\newblock In \emph{Proceedings of the 41st International Conference on Machine
  Learning}, 2024.
\newblock URL \url{https://arxiv.org/abs/2401.12926}.

\bibitem[Evans et~al.(2024)Evans, Parthasarathy, Merzic, and
  Henaff]{evans2024datacurationjointexample}
Talfan Evans, Nikhil Parthasarathy, Hamza Merzic, and Olivier~J. Henaff.
\newblock Data curation via joint example selection further accelerates
  multimodal learning.
\newblock In \emph{Advances in Neural Information Processing Systems}, 2024.
\newblock URL \url{https://arxiv.org/abs/2406.17711}.

\bibitem[Gebru et~al.(2021)Gebru, Morgenstern, Vecchione, Vaughan, Wallach,
  Daum{\'e}~III, and Crawford]{gebru2021datasheets}
Timnit Gebru, Jamie Morgenstern, Briana Vecchione, Jennifer~Wortman Vaughan,
  Hanna Wallach, Hal Daum{\'e}~III, and Kate Crawford.
\newblock Datasheets for datasets.
\newblock \emph{Communications of the ACM}, 64\penalty0 (12):\penalty0 86--92,
  2021.
\newblock \doi{10.1145/3458723}.
\newblock URL \url{https://doi.org/10.1145/3458723}.

\bibitem[Goyal et~al.(2024)Goyal, Maini, Lipton, Raghunathan, and
  Kolter]{goyal2024scaling}
Sachin Goyal, Pratyush Maini, Zachary~C. Lipton, Aditi Raghunathan, and J.~Zico
  Kolter.
\newblock Scaling laws for data filtering: Data curation cannot be compute
  agnostic.
\newblock In \emph{Proceedings of the IEEE/CVF Conference on Computer Vision
  and Pattern Recognition (CVPR)}, pages 22702--22711, 2024.
\newblock \doi{10.1109/CVPR52733.2024.02142}.
\newblock URL \url{https://doi.org/10.1109/CVPR52733.2024.02142}.

\bibitem[Grosse et~al.(2023)Grosse, Bae, Anil, Elhage, Tamkin, Tajdini,
  Steiner, Li, Durmus, Perez, Hubinger, Luko{\v s}i{\=u}t{\.e}, Nguyen, Joseph,
  McCandlish, Kaplan, and Bowman]{grosse2023studying}
Roger Grosse, Juhan Bae, Cem Anil, Nelson Elhage, Alex Tamkin, Amirhossein
  Tajdini, Benoit Steiner, Dustin Li, Esin Durmus, Ethan Perez, Evan Hubinger,
  Kamil{\.e} Luko{\v s}i{\=u}t{\.e}, Karina Nguyen, Nicholas Joseph, Sam
  McCandlish, Jared Kaplan, and Samuel~R. Bowman.
\newblock Studying large language model generalization with influence
  functions.
\newblock arXiv preprint arXiv:2308.03296, 2023.
\newblock URL \url{https://arxiv.org/abs/2308.03296}.

\bibitem[Gu et~al.(2025{\natexlab{a}})Gu, Tafjord, Kuehl, Haddad, Dodge, and
  Hajishirzi]{gu2025olmes}
Yuling Gu, Oyvind Tafjord, Bailey Kuehl, Dany Haddad, Jesse Dodge, and Hannaneh
  Hajishirzi.
\newblock {OLMES}: A standard for language model evaluations.
\newblock In \emph{Findings of the Association for Computational Linguistics:
  NAACL 2025}, pages 5020--5048. Association for Computational Linguistics,
  April 2025{\natexlab{a}}.
\newblock \doi{10.18653/v1/2025.findings-naacl.282}.
\newblock URL \url{https://aclanthology.org/2025.findings-naacl.282/}.

\bibitem[Gu et~al.(2025{\natexlab{b}})Gu, Dong, Wang, Hao, Dong, Wei, and
  Huang]{gu2025dataselectionoptimalcontrol}
Yuxian Gu, Li~Dong, Hongning Wang, Yaru Hao, Qingxiu Dong, Furu Wei, and Minlie
  Huang.
\newblock Data selection via optimal control for language models.
\newblock In \emph{International Conference on Learning Representations},
  2025{\natexlab{b}}.
\newblock URL \url{https://arxiv.org/abs/2410.07064}.

\bibitem[Guu et~al.(2023)Guu, Webson, Pavlick, Dixon, Tenney, and
  Bolukbasi]{guu2023simfluence}
Kelvin Guu, Albert Webson, Ellie Pavlick, Lucas Dixon, Ian Tenney, and Tolga
  Bolukbasi.
\newblock Simfluence: Modeling the influence of individual training examples by
  simulating training runs.
\newblock arXiv preprint arXiv:2303.08114, 2023.
\newblock URL \url{https://arxiv.org/abs/2303.08114}.

\bibitem[Heo et~al.(2026)Heo, Yun, Choi, Hwang, Ok, and
  Kim]{heo2026interactionawareinfluencefunctionsgroup}
Jaeseung Heo, Kyeongheung Yun, Youngbin Choi, Sehyun Hwang, Jungseul Ok, and
  Dongwoo Kim.
\newblock Interaction-aware influence functions for group attribution.
\newblock arXiv preprint arXiv:2605.15675, 2026.
\newblock URL \url{https://arxiv.org/abs/2605.15675}.

\bibitem[Ilyas et~al.(2022)Ilyas, Park, Engstrom, Leclerc, and
  Madry]{ilyas2022datamodels}
Andrew Ilyas, Sung~Min Park, Logan Engstrom, Guillaume Leclerc, and Aleksander
  Madry.
\newblock Datamodels: Understanding predictions with data and data with
  predictions.
\newblock In \emph{Proceedings of the 39th International Conference on Machine
  Learning}, volume 162 of \emph{Proceedings of Machine Learning Research},
  pages 9525--9587, 2022.
\newblock URL \url{https://proceedings.mlr.press/v162/ilyas22a.html}.

\bibitem[Jiang et~al.(2025)Jiang, Zhou, Feng, Malladi, and
  Kolter]{jiang2024adaptivedataoptimizationdynamic}
Yiding Jiang, Allan Zhou, Zhili Feng, Sadhika Malladi, and J.~Zico Kolter.
\newblock Adaptive data optimization: Dynamic sample selection with scaling
  laws.
\newblock In \emph{International Conference on Learning Representations}, 2025.
\newblock URL \url{https://arxiv.org/abs/2410.11820}.

\bibitem[Jiao et~al.(2025)Jiao, Pan, Xiao, Sheng, Jain, Zhao, Dasgupta, Ma, and
  Xiong]{jiao2025date}
Cathy Jiao, Yijun Pan, Emily Xiao, Daisy Sheng, Niket Jain, Hanzhang Zhao,
  Ishita Dasgupta, Jiaqi~W. Ma, and Chenyan Xiong.
\newblock {DATE-LM}: Benchmarking data attribution evaluation for large
  language models.
\newblock In \emph{Advances in Neural Information Processing Systems Datasets
  and Benchmarks Track}, volume~38, 2025.
\newblock \doi{10.52202/085713-5142}.
\newblock URL \url{https://arxiv.org/abs/2507.14630}.

\bibitem[Khaddaj et~al.(2025)Khaddaj, Engstrom, and
  Madry]{khaddaj2025smalltolargegeneralizationdatainfluences}
Alaa Khaddaj, Logan Engstrom, and Aleksander Madry.
\newblock Small-to-large generalization: Data influences models consistently
  across scale.
\newblock In \emph{International Conference on Learning Representations}, 2025.
\newblock URL \url{https://arxiv.org/abs/2505.16260}.

\bibitem[Ko et~al.(2025)Ko, Just, Fleming, Jin, and Jia]{ko2025probing}
Myeongseob Ko, Hoang~Anh Just, Charles Fleming, Ming Jin, and Ruoxi Jia.
\newblock Probing knowledge holes in unlearned {LLM}s.
\newblock In \emph{Advances in Neural Information Processing Systems},
  volume~38, 2025.
\newblock URL \url{https://arxiv.org/abs/2511.00030}.

\bibitem[Koh and Liang(2017)]{koh2017understanding}
Pang~Wei Koh and Percy Liang.
\newblock Understanding black-box predictions via influence functions.
\newblock In \emph{Proceedings of the 34th International Conference on Machine
  Learning}, volume~70 of \emph{Proceedings of Machine Learning Research},
  pages 1885--1894, 2017.
\newblock URL \url{https://proceedings.mlr.press/v70/koh17a.html}.

\bibitem[Lee et~al.(2026)Lee, Smith, Adam, and Hoogland]{lee2026influence}
Jin~Hwa Lee, Matthew Smith, Maxwell Adam, and Jesse Hoogland.
\newblock Influence dynamics and stagewise data attribution.
\newblock In \emph{International Conference on Learning Representations}, pages
  35716--35747, 2026.
\newblock URL \url{https://arxiv.org/abs/2510.12071}.

\bibitem[Lehalleur et~al.(2025)Lehalleur, Hoogland, Farrugia-Roberts, Wei,
  Oldenziel, Wang, Carroll, and Murfet]{lehalleur2025youarewhatyoueat}
Simon~Pepin Lehalleur, Jesse Hoogland, Matthew Farrugia-Roberts, Susan Wei,
  Alexander~Gietelink Oldenziel, George Wang, Liam Carroll, and Daniel Murfet.
\newblock You are what you eat: {AI} alignment requires understanding how data
  shapes structure and generalisation.
\newblock arXiv preprint arXiv:2502.05475, 2025.
\newblock URL \url{https://arxiv.org/abs/2502.05475}.

\bibitem[Li et~al.(2024)Li, Fang, Smyrnis, Ivgi, Jordan, Gadre, Bansal, Guha,
  Keh, Arora, Garg, Xin, Muennighoff, Heckel, Mercat, Chen, Gururangan,
  Wortsman, Albalak, Bitton, Nezhurina, Abbas, Hsieh, Ghosh, Gardner, Kilian,
  Zhang, Shao, Pratt, Sanyal, Ilharco, Daras, Marathe, Gokaslan, Zhang, Chandu,
  Nguyen, Vasiljevic, Kakade, Song, Sanghavi, Faghri, Oh, Zettlemoyer, Lo,
  El-Nouby, Pouransari, Toshev, Wang, Groeneveld, Soldaini, Koh, Jitsev,
  Kollar, Dimakis, Carmon, Dave, Schmidt, and Shankar]{li2024datacomplm}
Jeffrey Li, Alex Fang, Georgios Smyrnis, Maor Ivgi, Matt Jordan, Samir Gadre,
  Hritik Bansal, Etash Guha, Sedrick Keh, Kushal Arora, Saurabh Garg, Rui Xin,
  Niklas Muennighoff, Reinhard Heckel, Jean Mercat, Mayee Chen, Suchin
  Gururangan, Mitchell Wortsman, Alon Albalak, Yonatan Bitton, Marianna
  Nezhurina, Amro Abbas, Cheng-Yu Hsieh, Dhruba Ghosh, Josh Gardner, Maciej
  Kilian, Hanlin Zhang, Rulin Shao, Sarah Pratt, Sunny Sanyal, Gabriel Ilharco,
  Giannis Daras, Kalyani Marathe, Aaron Gokaslan, Jieyu Zhang, Khyathi Chandu,
  Thao Nguyen, Igor Vasiljevic, Sham Kakade, Shuran Song, Sujay Sanghavi,
  Fartash Faghri, Sewoong Oh, Luke Zettlemoyer, Kyle Lo, Alaaeldin El-Nouby,
  Hadi Pouransari, Alexander Toshev, Stephanie Wang, Dirk Groeneveld, Luca
  Soldaini, Pang~Wei Koh, Jenia Jitsev, Thomas Kollar, Alexandros~G. Dimakis,
  Yair Carmon, Achal Dave, Ludwig Schmidt, and Vaishaal Shankar.
\newblock {DataComp-LM}: In search of the next generation of training sets for
  language models.
\newblock In \emph{Advances in Neural Information Processing Systems Datasets
  and Benchmarks Track}, 2024.
\newblock URL \url{https://arxiv.org/abs/2406.11794}.

\bibitem[Lin et~al.(2024)Lin, Gou, Gong, Liu, Shen, Xu, Lin, Yang, Jiao, Duan,
  and Chen]{lin2024rho1}
Zhenghao Lin, Zhibin Gou, Yeyun Gong, Xiao Liu, Yelong Shen, Ruochen Xu, Chen
  Lin, Yujiu Yang, Jian Jiao, Nan Duan, and Weizhu Chen.
\newblock Not all tokens are what you need for pretraining.
\newblock In \emph{Advances in Neural Information Processing Systems},
  volume~37, pages 29029--29063, 2024.
\newblock \doi{10.52202/079017-0914}.
\newblock URL \url{https://arxiv.org/abs/2404.07965}.

\bibitem[Liu et~al.(2025{\natexlab{a}})Liu, Neubig, and
  Xiong]{liu2026midtrainingbridgespretrainingposttraining}
Emmy Liu, Graham Neubig, and Chenyan Xiong.
\newblock Midtraining bridges pretraining and posttraining distributions.
\newblock arXiv preprint arXiv:2510.14865, 2025{\natexlab{a}}.
\newblock URL \url{https://arxiv.org/abs/2510.14865}.

\bibitem[Liu et~al.(2025{\natexlab{b}})Liu, Blanton, Elazar, Min, Chen,
  Chheda-Kothary, Tran, Bischoff, Marsh, Schmitz, Trier, Sarnat, James,
  Borchardt, Kuehl, Cheng, Farley, Anderson, Albright, Schoenick, Soldaini,
  Groeneveld, Finney, Pang, Ramanujan, Lo, Zettlemoyer, Tsvetkov, Smith,
  Hajishirzi, Farhadi, and Koh]{liu2025olmotracetracinglanguagemodel}
Jiacheng Liu, Taylor Blanton, Yanai Elazar, Sewon Min, Yen-Sung Chen, Arnavi
  Chheda-Kothary, Huy Tran, Byron Bischoff, Eric Marsh, Michael Schmitz,
  Cassidy Trier, Aaron Sarnat, Jenna James, Jon Borchardt, Bailey Kuehl, Evie
  Yu-Yen Cheng, Karen Farley, Taira Anderson, David Albright, Carissa
  Schoenick, Luca Soldaini, Dirk Groeneveld, Rock~Yuren Finney, Ruochen Pang,
  Vikram Ramanujan, Kyle Lo, Luke Zettlemoyer, Yulia Tsvetkov, Noah~A. Smith,
  Hannaneh Hajishirzi, Ali Farhadi, and Pang~Wei Koh.
\newblock {OLMoTrace}: Tracing language model outputs back to trillions of
  training tokens.
\newblock In \emph{Proceedings of the 63rd Annual Meeting of the Association
  for Computational Linguistics (Volume 3: System Demonstrations)}, pages
  178--188, 2025{\natexlab{b}}.
\newblock \doi{10.18653/v1/2025.acl-demo.18}.
\newblock URL \url{https://doi.org/10.18653/v1/2025.acl-demo.18}.

\bibitem[Liu et~al.(2025{\natexlab{c}})Liu, Zheng, Muennighoff, Zeng, Dou,
  Pang, Jiang, and Lin]{liu2025regmixdatamixtureregression}
Qian Liu, Xiaosen Zheng, Niklas Muennighoff, Guangtao Zeng, Longxu Dou, Tianyu
  Pang, Jing Jiang, and Min Lin.
\newblock {RegMix}: Data mixture as regression for language model pre-training.
\newblock In \emph{International Conference on Learning Representations},
  2025{\natexlab{c}}.
\newblock URL \url{https://arxiv.org/abs/2407.01492}.

\bibitem[{\L}ucki et~al.(2025){\L}ucki, Wei, Huang, Henderson, Tram{\`e}r, and
  Rando]{lucki2025adversarial}
Jakub {\L}ucki, Boyi Wei, Yangsibo Huang, Peter Henderson, Florian Tram{\`e}r,
  and Javier Rando.
\newblock An adversarial perspective on machine unlearning for {AI} safety.
\newblock \emph{Transactions on Machine Learning Research}, 2025.
\newblock URL \url{https://arxiv.org/abs/2409.18025}.

\bibitem[Magnusson et~al.(2025)Magnusson, Tai, Bogin, Heineman, Hwang,
  Soldaini, Bhagia, Liu, Groeneveld, Tafjord, Smith, Koh, and
  Dodge]{magnusson2025datadecide}
Ian Magnusson, Nguyen Tai, Ben Bogin, David Heineman, Jena~D. Hwang, Luca
  Soldaini, Akshita Bhagia, Jiacheng Liu, Dirk Groeneveld, Oyvind Tafjord,
  Noah~A. Smith, Pang~Wei Koh, and Jesse Dodge.
\newblock {DataDecide}: How to predict best pretraining data with small
  experiments.
\newblock In \emph{Proceedings of the 42nd International Conference on Machine
  Learning}, 2025.
\newblock URL \url{https://arxiv.org/abs/2504.11393}.

\bibitem[Matlin et~al.(2026)Matlin, Chakraborty, Eom, Okamoto, Castilla,
  Jaburi, Deng, Min, Quirke, Biderman, and Riedl]{matlin2026capability}
Glenn Matlin, Chandreyi Chakraborty, Saehee Eom, Mika Okamoto, Rayan Castilla,
  Louis Jaburi, Alvin Deng, Taywon Min, Lucia Quirke, Stella Biderman, and Mark
  Riedl.
\newblock Capability provenance in language models: A case study in social
  reasoning.
\newblock In \emph{Conference on Language Modeling}, 2026.
\newblock URL \url{https://arxiv.org/abs/2606.19625}.

\bibitem[Mindermann et~al.(2022)Mindermann, Brauner, Razzak, Sharma, Kirsch,
  Xu, H{\"o}ltgen, Gomez, Morisot, Farquhar, and
  Gal]{mindermann2022prioritized}
S{\"o}ren Mindermann, Jan Brauner, Muhammed Razzak, Mrinank Sharma, Andreas
  Kirsch, Winnie Xu, Benedikt H{\"o}ltgen, Aidan~N. Gomez, Adrien Morisot,
  Sebastian Farquhar, and Yarin Gal.
\newblock Prioritized training on points that are learnable, worth learning,
  and not yet learnt.
\newblock In \emph{Proceedings of the 39th International Conference on Machine
  Learning}, 2022.
\newblock URL \url{https://arxiv.org/abs/2206.07137}.

\bibitem[Mohri et~al.(2026)Mohri, Duchi, and
  Hashimoto]{mohri2026bitterlessondatafiltering}
Christopher Mohri, John Duchi, and Tatsunori Hashimoto.
\newblock A bitter lesson for data filtering.
\newblock arXiv preprint arXiv:2605.19407, 2026.
\newblock URL \url{https://arxiv.org/abs/2605.19407}.

\bibitem[Muennighoff et~al.(2023)Muennighoff, Rush, Barak, Le~Scao, Tazi,
  Piktus, Pyysalo, Wolf, and Raffel]{muennighoff2023scaling}
Niklas Muennighoff, Alexander Rush, Boaz Barak, Teven Le~Scao, Nouamane Tazi,
  Aleksandra Piktus, Sampo Pyysalo, Thomas Wolf, and Colin~A. Raffel.
\newblock Scaling data-constrained language models.
\newblock In \emph{Advances in Neural Information Processing Systems},
  volume~36, pages 50358--50376, 2023.
\newblock URL \url{https://arxiv.org/abs/2305.16264}.

\bibitem[Nguyen et~al.(2024)Nguyen, Bertram, Kortukov, Song, and
  Oh]{nguyen2024towards}
Elisa Nguyen, Johannes Bertram, Evgenii Kortukov, Jean~Y. Song, and Seong~Joon
  Oh.
\newblock Towards user-focused research in training data attribution for
  human-centered explainable {AI}.
\newblock arXiv preprint arXiv:2409.16978, 2024.
\newblock URL \url{https://arxiv.org/abs/2409.16978}.

\bibitem[Nishida et~al.(2026)Nishida, Kiyomaru, Oda, Kodama, Liu, Kawahara,
  Miyao, M{\"u}ller-Eberstein, and Isonuma]{nishida2026measuring}
Yuto Nishida, Hirokazu Kiyomaru, Yusuke Oda, Takashi Kodama, Chaoran Liu,
  Daisuke Kawahara, Yusuke Miyao, Max M{\"u}ller-Eberstein, and Masaru Isonuma.
\newblock Measuring task-agnostic training data influence across language model
  pretraining.
\newblock In \emph{Conference on Language Modeling}, 2026.
\newblock URL \url{https://arxiv.org/abs/2608.13515}.

\bibitem[Orgad et~al.(2026)Orgad, Barez, Haklay, Lee, Mosbach, Reusch, Saphra,
  Wallace, Wiegreffe, Wong, Tenney, and Geva]{orgad2026actionable}
Hadas Orgad, Fazl Barez, Tal Haklay, Isabelle Lee, Marius Mosbach, Anja Reusch,
  Naomi Saphra, Byron Wallace, Sarah Wiegreffe, Eric Wong, Ian Tenney, and Mor
  Geva.
\newblock Interpretability can be actionable.
\newblock In \emph{Proceedings of the 43rd International Conference on Machine
  Learning}, 2026.
\newblock URL \url{https://arxiv.org/abs/2605.11161}.

\bibitem[Park et~al.(2023)Park, Georgiev, Ilyas, Leclerc, and
  Madry]{park2023trak}
Sung~Min Park, Kristian Georgiev, Andrew Ilyas, Guillaume Leclerc, and
  Aleksander Madry.
\newblock {TRAK}: Attributing model behavior at scale.
\newblock In \emph{Proceedings of the 40th International Conference on Machine
  Learning}, volume 202 of \emph{Proceedings of Machine Learning Research},
  pages 27074--27113, 2023.
\newblock URL \url{https://proceedings.mlr.press/v202/park23c.html}.

\bibitem[Penedo et~al.(2024)Penedo, Kydl{\'\i}{\v{c}}ek, Ben~Allal, Lozhkov,
  Mitchell, Raffel, Von~Werra, and Wolf]{penedo2024fineweb}
Guilherme Penedo, Hynek Kydl{\'\i}{\v{c}}ek, Loubna Ben~Allal, Anton Lozhkov,
  Margaret Mitchell, Colin Raffel, Leandro Von~Werra, and Thomas Wolf.
\newblock The {FineWeb} datasets: Decanting the web for the finest text data at
  scale.
\newblock In \emph{Advances in Neural Information Processing Systems},
  volume~37, pages 30811--30849, 2024.
\newblock \doi{10.52202/079017-0970}.
\newblock URL \url{https://arxiv.org/abs/2406.17557}.

\bibitem[Pruthi et~al.(2020)Pruthi, Liu, Kale, and
  Sundararajan]{pruthi2020estimating}
Garima Pruthi, Frederick Liu, Satyen Kale, and Mukund Sundararajan.
\newblock Estimating training data influence by tracing gradient descent.
\newblock In \emph{Advances in Neural Information Processing Systems},
  volume~33, pages 19920--19930, 2020.
\newblock URL \url{https://arxiv.org/abs/2002.08484}.

\bibitem[Rathi and Radford(2026)]{rathi2026shapingcapabilitiestokenleveldata}
Neil Rathi and Alec Radford.
\newblock Shaping capabilities with token-level data filtering.
\newblock arXiv preprint arXiv:2601.21571, 2026.
\newblock URL \url{https://arxiv.org/abs/2601.21571}.

\bibitem[Saada et~al.(2026)Saada, Bethune, Klein, Grangier, Cuturi, and
  Ablin]{saada2026removingnoisefindinggold}
Thiziri~Nait Saada, Louis Bethune, Michal Klein, David Grangier, Marco Cuturi,
  and Pierre Ablin.
\newblock Removing noise, not finding gold: Quality filtering for large-scale
  pretraining.
\newblock In \emph{Proceedings of the 43rd International Conference on Machine
  Learning}, 2026.
\newblock URL \url{https://arxiv.org/abs/2510.00866}.

\bibitem[Sam et~al.(2025)Sam, Chakrabarti, Rostamizadeh, Ramalingam, Citovsky,
  and Kumar]{sam2025analyzingsimilaritymetricsdata}
Dylan Sam, Ayan Chakrabarti, Afshin Rostamizadeh, Srikumar Ramalingam, Gui
  Citovsky, and Sanjiv Kumar.
\newblock Analyzing similarity metrics for data selection for language model
  pretraining.
\newblock In \emph{Advances in Neural Information Processing Systems},
  volume~38, 2025.
\newblock \doi{10.52202/085713-4543}.
\newblock URL \url{https://arxiv.org/abs/2502.02494}.

\bibitem[Schoenegger et~al.(2025)Schoenegger, Thoma, Blevins, and
  Roth]{babylm2025influence}
Loris Schoenegger, Lukas Thoma, Terra Blevins, and Benjamin Roth.
\newblock Influence-driven curriculum learning for pre-training on limited
  data.
\newblock In \emph{Proceedings of the First BabyLM Workshop}, pages 356--379,
  Suzhou, China, November 2025. Association for Computational Linguistics.
\newblock \doi{10.18653/v1/2025.babylm-main.26}.
\newblock URL \url{https://aclanthology.org/2025.babylm-main.26/}.

\bibitem[Shum et~al.(2025)Shum, Huang, Zou, Ding, Liao, Chen, Liu, and
  He]{shum2025predictivedataselectiondata}
Kashun Shum, Yuzhen Huang, Hongjian Zou, Qi~Ding, Yixuan Liao, Xiaoxin Chen,
  Qian Liu, and Junxian He.
\newblock Predictive data selection: The data that predicts is the data that
  teaches.
\newblock In \emph{Proceedings of the 42nd International Conference on Machine
  Learning}, 2025.
\newblock URL \url{https://arxiv.org/abs/2503.00808}.

\bibitem[Thrush et~al.(2025)Thrush, Potts, and
  Hashimoto]{thrush2025improvingpretrainingdatausing}
Tristan Thrush, Christopher Potts, and Tatsunori Hashimoto.
\newblock Improving pretraining data using perplexity correlations.
\newblock In \emph{International Conference on Learning Representations}, 2025.
\newblock URL \url{https://arxiv.org/abs/2409.05816}.

\bibitem[Wang et~al.(2024)Wang, Song, Zou, Mittal, and
  Jia]{wang2024capturingtemporaldependencetraining}
Jiachen~T. Wang, Dawn Song, James Zou, Prateek Mittal, and Ruoxi Jia.
\newblock Capturing the temporal dependence of training data influence.
\newblock arXiv preprint arXiv:2412.09538, 2024.
\newblock URL \url{https://arxiv.org/abs/2412.09538}.

\bibitem[Wang et~al.(2025{\natexlab{a}})Wang, Mittal, Song, and
  Jia]{wang2025inrunshapley}
Jiachen~T. Wang, Prateek Mittal, Dawn Song, and Ruoxi Jia.
\newblock Data {S}hapley in one training run.
\newblock In \emph{International Conference on Learning Representations},
  2025{\natexlab{a}}.
\newblock URL \url{https://arxiv.org/abs/2406.11011}.

\bibitem[Wang et~al.(2026{\natexlab{a}})Wang, Wu, Lyu, Zou, Song, Jia, and
  Mittal]{wang2026smalltrainingruns}
Jiachen~T. Wang, Tong Wu, Kaifeng Lyu, James Zou, Dawn Song, Ruoxi Jia, and
  Prateek Mittal.
\newblock Can small training runs reliably guide data curation? {R}ethinking
  proxy-model practice.
\newblock In \emph{International Conference on Learning Representations},
  2026{\natexlab{a}}.
\newblock URL \url{https://arxiv.org/abs/2512.24503}.

\bibitem[Wang et~al.(2026{\natexlab{b}})Wang, Woisetschl\"ager, Jacobsen, and
  Ji]{wang2026positionletsdevelopdata}
Shiqiang Wang, Herbert Woisetschl\"ager, Hans~Arno Jacobsen, and Mingyue Ji.
\newblock Position: Let's develop data probes to fundamentally understand how
  data affects {LLM} performance.
\newblock In \emph{Proceedings of the 43rd International Conference on Machine
  Learning}, 2026{\natexlab{b}}.
\newblock URL \url{https://arxiv.org/abs/2605.18801}.

\bibitem[Wang et~al.(2025{\natexlab{b}})Wang, Antoniades, Elazar, Amayuelas,
  Albalak, Zhang, and Wang]{wang2025generalizationvsmemorizationtracing}
Xinyi Wang, Antonis Antoniades, Yanai Elazar, Alfonso Amayuelas, Alon Albalak,
  Kexun Zhang, and William~Yang Wang.
\newblock Generalization v.s. memorization: Tracing language models'
  capabilities back to pretraining data.
\newblock In \emph{International Conference on Learning Representations},
  2025{\natexlab{b}}.
\newblock URL \url{https://arxiv.org/abs/2407.14985}.

\bibitem[Wang et~al.(2026{\natexlab{c}})Wang, Liu, Liu, Guo, Deng, Wu, Zhou,
  Zhou, and Wang]{wang2025tikmixdatainfluencedynamic}
Yifan Wang, Binbin Liu, Fengze Liu, Yuanfan Guo, Jiyao Deng, Xuecheng Wu,
  Weidong Zhou, Xiaohuan Zhou, and Taifeng Wang.
\newblock {TiKMiX}: Efficient semi-dynamic data mixture via data influence for
  {LLM} pre-training.
\newblock In \emph{Proceedings of the 64th Annual Meeting of the Association
  for Computational Linguistics (Volume 1: Long Papers)}, pages 5777--5793,
  2026{\natexlab{c}}.
\newblock \doi{10.18653/v1/2026.acl-long.261}.
\newblock URL \url{https://doi.org/10.18653/v1/2026.acl-long.261}.

\bibitem[Wettig et~al.(2024)Wettig, Gupta, Malik, and Chen]{wettig2024qurating}
Alexander Wettig, Aatmik Gupta, Saumya Malik, and Danqi Chen.
\newblock {QuRating}: Selecting high-quality data for training language models.
\newblock In \emph{Proceedings of the 41st International Conference on Machine
  Learning}, 2024.
\newblock URL \url{https://arxiv.org/abs/2402.09739}.

\bibitem[Wettig et~al.(2025)Wettig, Lo, Min, Hajishirzi, Chen, and
  Soldaini]{wettig2025weborganizer}
Alexander Wettig, Kyle Lo, Sewon Min, Hannaneh Hajishirzi, Danqi Chen, and Luca
  Soldaini.
\newblock Organize the web: Constructing domains enhances pre-training data
  curation.
\newblock In \emph{Proceedings of the 42nd International Conference on Machine
  Learning}, 2025.
\newblock URL \url{https://arxiv.org/abs/2502.10341}.

\bibitem[Worledge et~al.(2024)Worledge, Shen, Meister, Winston, and
  Guestrin]{worledge2023unifying}
Theodora Worledge, Judy~Hanwen Shen, Nicole Meister, Caleb Winston, and Carlos
  Guestrin.
\newblock Unifying corroborative and contributive attributions in large
  language models.
\newblock In \emph{2024 IEEE Conference on Secure and Trustworthy Machine
  Learning (SaTML)}, pages 665--683, 2024.
\newblock \doi{10.1109/SaTML59370.2024.00039}.
\newblock URL \url{https://doi.org/10.1109/SaTML59370.2024.00039}.

\bibitem[Xia et~al.(2024)Xia, Malladi, Gururangan, Arora, and
  Chen]{xia2024less}
Mengzhou Xia, Sadhika Malladi, Suchin Gururangan, Sanjeev Arora, and Danqi
  Chen.
\newblock {LESS}: Selecting influential data for targeted instruction tuning.
\newblock In \emph{Proceedings of the 41st International Conference on Machine
  Learning}, 2024.
\newblock URL \url{https://arxiv.org/abs/2402.04333}.

\bibitem[Xie et~al.(2023{\natexlab{a}})Xie, Pham, Dong, Du, Liu, Lu, Liang, Le,
  Ma, and Yu]{xie2023doremi}
Sang~Michael Xie, Hieu Pham, Xuanyi Dong, Nan Du, Hanxiao Liu, Yifeng Lu, Percy
  Liang, Quoc~V. Le, Tengyu Ma, and Adams~Wei Yu.
\newblock {DoReMi}: Optimizing data mixtures speeds up language model
  pretraining.
\newblock In \emph{Advances in Neural Information Processing Systems},
  volume~36, pages 69798--69818, 2023{\natexlab{a}}.
\newblock URL \url{https://arxiv.org/abs/2305.10429}.

\bibitem[Xie et~al.(2023{\natexlab{b}})Xie, Santurkar, Ma, and
  Liang]{xie2023dsir}
Sang~Michael Xie, Shibani Santurkar, Tengyu Ma, and Percy Liang.
\newblock Data selection for language models via importance resampling.
\newblock In \emph{Advances in Neural Information Processing Systems},
  volume~36, pages 34201--34227, 2023{\natexlab{b}}.
\newblock URL \url{https://arxiv.org/abs/2302.03169}.

\bibitem[Yu et~al.(2024)Yu, Das, and Xiong]{yu2024mates}
Zichun Yu, Spandan Das, and Chenyan Xiong.
\newblock {MATES}: Model-aware data selection for efficient pretraining with
  data influence models.
\newblock In \emph{Advances in Neural Information Processing Systems}, 2024.
\newblock URL \url{https://arxiv.org/abs/2406.06046}.

\bibitem[Yu et~al.(2025)Yu, Peng, Lei, Overwijk, Yih, and
  Xiong]{yu2025groupmates}
Zichun Yu, Fei Peng, Jie Lei, Arnold Overwijk, {Wen-tau} Yih, and Chenyan
  Xiong.
\newblock Group-level data selection for efficient pretraining.
\newblock In \emph{Advances in Neural Information Processing Systems}, 2025.
\newblock \doi{10.52202/085713-5176}.
\newblock URL \url{https://arxiv.org/abs/2502.14709}.

\bibitem[Zhang et~al.(2026)Zhang, Mohamed, Abdine, Shang, and
  Vazirgiannis]{zhang2026beyond}
Yang Zhang, Amr Mohamed, Hadi Abdine, Guokan Shang, and Michalis Vazirgiannis.
\newblock Beyond random sampling: Efficient language model pretraining via
  curriculum learning.
\newblock In Vera Demberg, Kentaro Inui, and Llu{\'i}s Marquez, editors,
  \emph{Proceedings of the 19th Conference of the European Chapter of the
  Association for Computational Linguistics (Volume 1: Long Papers)}, pages
  5776--5794, Rabat, Morocco, March 2026. Association for Computational
  Linguistics.
\newblock ISBN 979-8-89176-380-7.
\newblock \doi{10.18653/v1/2026.eacl-long.271}.
\newblock URL \url{https://aclanthology.org/2026.eacl-long.271/}.

\end{thebibliography}

\clearpage
\appendix

\section{What attribution evidence supports}
\label{app:claim-ladder}

The value of an attribution depends on the decision it informs. Identifying influential documents, explaining what they share, and choosing what to train on next are distinct questions, and evidence for one does not settle the others. This appendix develops the five qualifying considerations introduced alongside Table~\ref{tab:contract} in Section~\ref{sec:evidence-contract} and maps prior systems onto its rows.

\paragraph{Interpreting a source is a scientific step.}
Structural similarity compares discourse organization; lexical overlap compares direct textual similarity. Either can suggest why attributed documents matter, but a topic, source identity, or repeated phrase may support competing explanations. Semantic coherence in an external representation does not establish that the language model whose provenance is being studied represents the same distinctions internally. A grouping should therefore help formulate and distinguish hypotheses, rather than settle the explanation by naming it.

Section~\ref{sec:control-loop} notes that a broad category can conceal opposing effects and that summed document scores do not give a group effect. Data outside a taxonomy can also be informative: text with an identifiable subject absent from the scheme reveals a missing category, whereas text too corrupted or incomplete to classify exposes a data-quality failure. Neither status alone establishes the text's contribution to learning, because a curation rule acts on document properties while an attribution estimate answers a question about a specific model and target.

\paragraph{From explanation to a useful decision.}
An attribution score identifies a candidate source of support; interpreting its documents suggests a hypothesis about what that support consists of. A data intervention tests the resulting decision. Removal, replacement, and reweighting differ, so the replacement data and resource budget are part of the claim. Causal claims require comparisons that isolate the stated data change. A successful trial can justify a larger test without establishing that the policy transfers to every scale or training stage.

The decision may reweight a training mixture directly or induce a reusable criterion for selecting new data. The latter requires held-out corpus evaluation with discovery cost charged. Matched comparisons with cheaper selectors test whether attribution improves the decision, while separating controls distinguish the explanations behind that gain \citep{orgad2026actionable}. When an intervention fails, structured records of which row in Table~\ref{tab:contract} broke keep future runs from repeating the mistake.

\paragraph{Robustness and generalization qualify the claim.}
Attribution reports must decouple influence signs, magnitudes, and relative rank orderings, as rank correlations can obscure sign reversals and near-zero effects. Sensitivity checks can vary queries, phrasing, checkpoint selection, and estimator assumptions. Disagreement can reveal approximation error or different counterfactual questions; change across checkpoints may also reflect learning. General policies need evidence across relevant runs, while a useful local claim may remain tied to one setting. A relationship that holds on related probes need not hold after a change in training stage, corpus, or model, so the intended scope determines the relevant transfer question.

\paragraph{Selective improvement is a joint outcome.}
A capability gain must be evaluated alongside changes in adjacent capabilities, safety, and general language modeling. A policy is selective only relative to pre-declared bounds on those off-target metrics; improving the target metric alone does not establish an acceptable tradeoff. Task accuracy, model calibration, and uncertainty about an intervention's causal effect are distinct quantities: confident outputs do not imply that reweighting their support will help, and noisy influence estimates do not imply model miscalibration.

\paragraph{Existing systems and the evidence they supply.}
Several systems already instantiate parts of the loop, and each supplies evidence for a specific row of Table~\ref{tab:contract}. Influence-driven mixture controllers \citep{yu2024mates,wang2025tikmixdatainfluencedynamic,chang2026automixercheckpointartifactsautomatic}, online mixing from per-domain scaling laws \citep{jiang2024adaptivedataoptimizationdynamic}, and optimal-control selection \citep{gu2025dataselectionoptimalcontrol} report improvements at matched compute in the scales and mixtures tested, with losses on some policies. They answer the intervention question for a chosen policy; they do not by themselves explain which data properties produced the gain. Trajectory-aware estimators \citep{wang2024capturingtemporaldependencetraining,bae2024trainingdataattributionapproximate} and cross-scale consistency of influence \citep{khaddaj2025smalltolargegeneralizationdatainfluences} address the robustness and transfer questions, while joint batch selection \citep{evans2024datacurationjointexample} and interaction-aware group influence \citep{heo2026interactionawareinfluencefunctionsgroup} address aggregation. Fact tracing \citep{Akyurek_2022}, cross-lingual sharing \citep{Choenni_2023}, and trillion-token verbatim tracing \citep{liu2025olmotracetracinglanguagemodel} supply descriptive evidence about what models draw on, at different costs and with different counterfactual meaning. Forgetting dynamics \citep{chang2024largelanguagemodelsacquire} bound how long an observed contribution persists. Reading these results through the table clarifies what remains open: a validated explanation of why a selected region helped, and evidence that a criterion learned from one run improves the next.

\paragraph{Mechanistic understanding has a distinct payoff.}
A mechanistic account answers an explanatory question and does not by itself establish transfer or decision value. Targeted data interventions during pretraining have changed the emergence of interpretable attention heads, with concurrent changes in in-context learning in the reported setting \citep{chen2026mechanistic}. Such a result remains scoped to the model, data, and intervention used. Mechanistic evidence can deepen an account of how data supports an ability, while a useful curation decision can be tested before that account is complete.

\paragraph{Illustrative reference protocol for Section~\ref{sec:research-agenda}.}
To show how the matched-cost comparison in Section~\ref{sec:research-agenda} can be operationalized in a single open testbed, Table~\ref{tab:protocol} provides an illustrative reference instantiation using released corpora, sub-billion to 1B model scales, and standardized evaluation suites. Concrete token budgets, checkpoint counts, and non-inferiority tolerances scale with the testbed, whereas pre-declaring all seven protocol components separates a discovery corpus $\mathcal{D}_{\text{disc}}$ (where attribution is computed and a candidate data property is extracted) from a disjoint transfer corpus $\mathcal{D}_{\text{hold}}$ (where the resulting criterion competes against cheaper selectors without further attribution).

\begin{table}[!htbp]
  \caption{Illustrative single-testbed protocol instantiating the matched-cost comparison of Section~\ref{sec:research-agenda}.}
  \label{tab:protocol}
  \centering
  \footnotesize
  \renewcommand{\arraystretch}{1.08}
  \begin{tabular}{@{}>{\raggedright\arraybackslash}p{0.22\linewidth}>{\raggedright\arraybackslash}p{0.73\linewidth}@{}}
    \toprule
    \textbf{Protocol component} & \textbf{Illustrative specification} \\
    \midrule
    \textbf{1. Corpora and split} & \emph{Discovery pool ($\mathcal{D}_{\text{disc}}$):} 10B-token stratified slice of DCLM-Baseline or Dolma labeled with WebOrganizer $24\times 24$ topic-by-format bins \citep{li2024datacomplm,wettig2025weborganizer}. \emph{Held-out transfer pool ($\mathcal{D}_{\text{hold}}$):} disjoint 20B-token DCLM-Pool slice untouched during discovery. \\
    \textbf{2. Model scale and checkpoints} & \emph{Discovery run:} 410M- or 1B-parameter decoder-only model trained for 10B tokens on $\mathcal{D}_{\text{disc}}$, saving 10 evenly spaced checkpoints with projected gradients ($d = 2048$--$4096$) indexed via TRAK or EK-FAC \citep{park2023trak,coalson2025ifguide}. \emph{Transfer run:} 1B-parameter model trained from scratch for 10B selected tokens from $\mathcal{D}_{\text{hold}}$ (plus a mid-training branch at 50\% budget for stage-conditional tests). \\
    \textbf{3. Criterion extraction} & Contrast top- and bottom-attributed regions or token clusters on $\mathcal{D}_{\text{disc}}$ across competing hypotheses (discourse structure, lexical overlap, source metadata), screen candidate edits via bounded end-of-training ablations \citep{blakeney2024doesdatasparkjoy} or unlearning, and compile the surviving property into a standalone classifier or token filter \citep{wettig2024qurating,rathi2026shapingcapabilitiestokenleveldata} that runs on $\mathcal{D}_{\text{hold}}$ without gradients. \\
    \textbf{4. Baseline selectors} & (a) Small-model perplexity pruning and perplexity correlations \citep{ankner2025perplexed,thrush2025improvingpretrainingdatausing}; (b) reference-corpus classifier and quality raters \citep{penedo2024fineweb,wettig2024qurating}; (c) embedding-centroid similarity to discovery queries; and (d) domain reweighting or online mixing \citep{xie2023doremi,albalak2023efficient}. \\
    \textbf{5. Matched total cost ($C_{\text{total}}$)} & Let $C_{\text{disc}} = C_{\text{index}} + C_{\text{query}} + C_{\text{screen}} + C_{\text{distill}}$ be all accelerator FLOPs (or GPU-hours) spent on $\mathcal{D}_{\text{disc}}$ beyond the base run, paired with an explicit cap on human inspection (e.g., $\le 8$ analyst-hours, or 0 hours when extraction uses automated lexical and discourse contrasts). Baseline arms pay only their selection cost $C_{\text{sel}}$ on $\mathcal{D}_{\text{hold}}$ and invest the difference $C_{\text{disc}} - C_{\text{sel}}$ into extra training tokens or mixture-weight sweeps on $\mathcal{D}_{\text{hold}}$ so every arm expends $C_{\text{total}} = C_{\text{disc}} + C_{\text{train}}$. \\
    \textbf{6. Probe suite and guardrails} & Evaluated under OLMES \citep{gu2025olmes} with above-chance admission floors. \emph{Discovery queries:} e.g., SocialIQA (social reasoning), GSM8K (math), or RealToxicityPrompts (safety). \emph{Disjoint held-out target probes:} unseen tasks and phrasings testing the same construct (e.g., social pragmatics, MATH-elementary, ToxiGen). \emph{Off-target guardrails:} MMLU macro-average, ARC-Challenge, HellaSwag, and Paloma validation perplexity. \\
    \textbf{7. Seeds and decision rule} & $S \ge 3$ independent pretraining seeds per arm on $\mathcal{D}_{\text{hold}}$ with paired bootstrap confidence intervals over evaluation items. The attribution-derived criterion succeeds if it improves the held-out target suite over the best compute-matched baseline ($p < 0.05$) while keeping every guardrail metric within a pre-declared non-inferiority bound (e.g., $\le 0.5$ percentage-point drop on off-target accuracy and $\le 1\%$ relative perplexity increase). Failing to beat the compute-matched baseline across two target constructs refutes the position at that scale. \\
    \bottomrule
  \end{tabular}
\end{table}
\clearpage

\section{Author contributions}
\label{app:contributions}

The matrix below records each author's roles under the NISO CRediT (Contributor Roles Taxonomy) standard, listed in byline order.

{%
\definecolor{creditFill}{HTML}{0072B2}%
\definecolor{creditStroke}{HTML}{004C77}%
\definecolor{creditEmpty}{HTML}{B8B8B8}%
\newcommand{\crY}{\tikz[baseline=-0.55ex]\filldraw[fill=creditFill, draw=creditStroke, line width=0.35pt] (0,0) circle (2.3pt);}%
\newcommand{\crN}{\tikz[baseline=-0.55ex]\fill[creditEmpty] (0,0) circle (0.85pt);}%
\newcommand{\crH}[1]{\multicolumn{1}{c}{\rotatebox{90}{\scriptsize #1}}}%
\begin{center}
\footnotesize
\setlength{\tabcolsep}{0pt}
\renewcommand{\arraystretch}{1.20}
\begin{tabular}{@{}l@{\hspace{1.2em}}*{14}{>{\centering\arraybackslash}p{1.95em}}@{}}
  & \crH{Conceptualization} & \crH{Data curation} & \crH{Formal analysis}
  & \crH{Funding acquisition} & \crH{Investigation} & \crH{Methodology}
  & \crH{Project administration} & \crH{Resources} & \crH{Software}
  & \crH{Supervision} & \crH{Validation} & \crH{Visualization}
  & \crH{Writing -- original draft} & \crH{Writing -- review \& editing} \\
\addlinespace[3pt]
\midrule
Glenn Matlin             & \crY & \crN & \crN & \crN & \crY & \crY & \crY & \crY & \crN & \crY & \crY & \crY & \crY & \crY \\
Chandreyi Chakraborty    & \crY & \crN & \crN & \crN & \crY & \crN & \crN & \crN & \crN & \crN & \crY & \crN & \crY & \crY \\
Mohammed Rehan Parwani   & \crY & \crN & \crY & \crN & \crN & \crY & \crN & \crN & \crN & \crN & \crY & \crN & \crY & \crY \\
Saehee Eom               & \crN & \crN & \crY & \crN & \crN & \crY & \crN & \crN & \crN & \crN & \crN & \crN & \crY & \crY \\
Anthony Wen-Ming Zang    & \crY & \crN & \crN & \crN & \crN & \crN & \crY & \crN & \crN & \crN & \crN & \crN & \crY & \crY \\
Akhil Theerthala         & \crY & \crN & \crN & \crN & \crY & \crY & \crN & \crN & \crN & \crN & \crN & \crN & \crY & \crN \\
Arjun Chatterjee         & \crY & \crN & \crY & \crN & \crY & \crY & \crN & \crN & \crN & \crN & \crY & \crY & \crY & \crY \\
Tejas Khandwekar         & \crN & \crN & \crN & \crN & \crY & \crY & \crN & \crN & \crN & \crN & \crN & \crY & \crY & \crY \\
Brandon Michaels         & \crY & \crN & \crN & \crN & \crY & \crY & \crN & \crN & \crN & \crN & \crN & \crN & \crY & \crY \\
Alvin Deng               & \crN & \crN & \crN & \crN & \crY & \crY & \crN & \crN & \crN & \crN & \crY & \crN & \crN & \crY \\
Mark Riedl               & \crN & \crN & \crN & \crN & \crN & \crN & \crN & \crN & \crN & \crY & \crN & \crN & \crN & \crN \\
\bottomrule
\end{tabular}
\end{center}
}

\end{document}